\documentclass[letterpaper, 10 pt, conference]{ieeeconf}  

\IEEEoverridecommandlockouts                              

\usepackage{graphicx} 
\usepackage{amsmath} 
\usepackage{multirow}
\usepackage{graphicx}
\usepackage{setspace}
\usepackage{tikz}
\usepackage{float}
\usepackage{amsmath}
\usepackage[ruled]{algorithm2e}
\usepackage{listings}
\usepackage{multirow} 
\usepackage{rotating}
\usepackage{color, colortbl}
\usepackage{pdfpages}
\usepackage{subcaption}
\usepackage{algorithmicx,algpseudocode}

\usepackage{url} 

\algnewcommand\algorithmicswitch{\textbf{switch}}
\algnewcommand\algorithmiccase{\textbf{case}}
\algnewcommand\algorithmicassert{\texttt{assert}}
\algnewcommand\Assert[1]{\State \algorithmicassert(#1)}%
\algdef{SE}[SWITCH]{Switch}{EndSwitch}[1]{\algorithmicswitch\ #1\ \algorithmicdo}{\algorithmicend\ \algorithmicswitch}%
\algdef{SE}[CASE]{Case}{EndCase}[1]{\algorithmiccase\ #1}{\algorithmicend\ \algorithmiccase}%
\algtext*{EndSwitch}%
\algtext*{EndCase}%

\title{\LARGE \bf
Multimodal feedback for active robot-object interaction
}

\author{Luis Contreras$^{1}$, Hiroki Yokoyama$^{2}$, and Hiroyuki Okada$^{3}$
\thanks{The authors are with the Advance Intelligence and Robotics Research Center at Tamagawa University, Japan.}%
\thanks{$^{1}${\tt\small contreras.luis@lab.tamagawa.ac.jp}}%
\thanks{$^{2}${\tt\small hiroki.yokoyama@okadanet.org}}%
\thanks{$^{3}${\tt\small h.okada@eng.tamagawa.ac.jp}}%
}

\begin{document}

\maketitle
\thispagestyle{empty}
\pagestyle{empty}

\begin{abstract}

In this work, we present a multimodal system for active robot-object interaction using laser-based SLAM, RGBD images, and contact sensors. In the object manipulation task, the robot adjusts its initial pose with respect to obstacles and target objects through RGBD data so it can perform object grasping in different configuration spaces while avoiding collisions, and updates the information related to the last steps of the manipulation process using the contact sensors in its hand. We perform a series of experiment to evaluate the performance of the proposed system following the the RoboCup2018 international competition regulations. We compare our approach with a number of baselines, namely a no-feedback method and visual-only and tactile-only feedback methods, where our proposed visual-and-tactile feedback method performs best.

\end{abstract}

\section{Introduction and Related Work}
\label{sec:introduction}

Autonomous service robots need several skills, such as motion planning in dynamic environments, object recognition and manipulation, human-robot interaction, and real-time awareness. These skills have been continuously evaluated in domestic environments and put into practice in competitions such as Robocup@Home \cite{wisspeintner:2009} and RoCKIn@Home \cite{amigoni:2015}, where the difficulty of the tasks increases each year to challenge the state-of-the-art technique. Nowadays, the tasks a service robot has to perform involves the integration of many skills, as individual systems can't provide all the information required to solve a given problem. One approach to system integration is through active perception, where the robot has to actively decide what to sense depending on the input sensors and the task at hand, and then act accordingly to the information received.

Such approaches have been proven useful in the navigation \cite{contreras:2017} and path planning tasks \cite{savage:2018} as well as for active multi-view object recognition \cite{potthast:2016}, where the authors make an intelligent feature sampling that performs well in a given task while decreasing computational time. Furthermore, \cite{potthast:2016} proposes an active view planning system to integrate new observations and deciding where to move the sensor next. However, when the task involves robot interaction with the scene, such as navigation without collisions or grasping objects without dropping them, current techniques favor mechanic and algorithmic precision rather than active reasoning \cite{cadena:2016} \cite{bajcsy:2018}.

Moreover, an important restriction comes from the mapping process itself. In the case of the Toyota Human Support Robot (HSR) \cite{hashimoto:2013}, it uses the Robot Operating System (ROS) \cite{quigley:2009} navigation stack, that includes path planning, obstacle avoidance, mapping and localization. In the mapping process, it generates an occupancy grid from a series of 2D laser readings, and differentiates between empty, occupied, and unknown space. As the sensor readings come from 2D scans at a fixed height, objects are partially mapped (e.g. only the legs from tables or chairs are added to the map) and the volumetric information is lost. 

In the object manipulation task, after the object detection and recognition steps, the robot has to plan the grasping trajectory such as it avoids collisions, and therefore the robot pose is limited by the arm's configuration space (i.e. all the possible poses the arm may take at a given robot position, considering the obstacles in the range). Localisation systems are able to update their map dynamically when changes are detected, and accurately pose the robot in a safe position to grasp an object. However, when changes are small or they happen after the robot reached the destination, an active perception system is desired to update its knowledge and perform the requested task successfully.

In this work, we present our active robot-object interaction system that uses multimodal feedback in the object manipulation task. More specifically, we present our solution for the RoboCup2018 Procter and Gamble (TM) Dishwasher Challenge \cite{matamoros:2018} where a service robot has to clean a table and place all dishes into a dishwasher. The dishwasher door is closed and the robot has to open it by manipulating the door handle. In the following sections we will present the details of our implementation.


\section{Multimodal robot-object interaction} \label{sect:robot}

\begin{figure}[ht]
	\centering
	\begin{subfigure}{.225\textwidth}
  		\centering
  		\includegraphics[width=1\textwidth]{{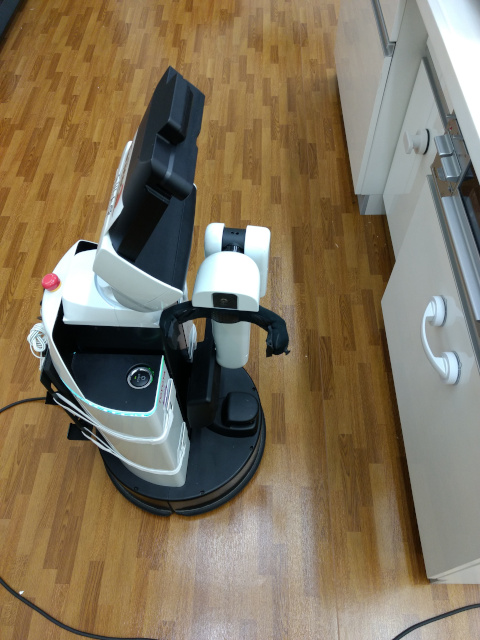}}
  		\caption{}
  		\label{fig:robotlateral}
	\end{subfigure}
	\begin{subfigure}{.225\textwidth}
  		\centering
  		\includegraphics[width=1\textwidth]{{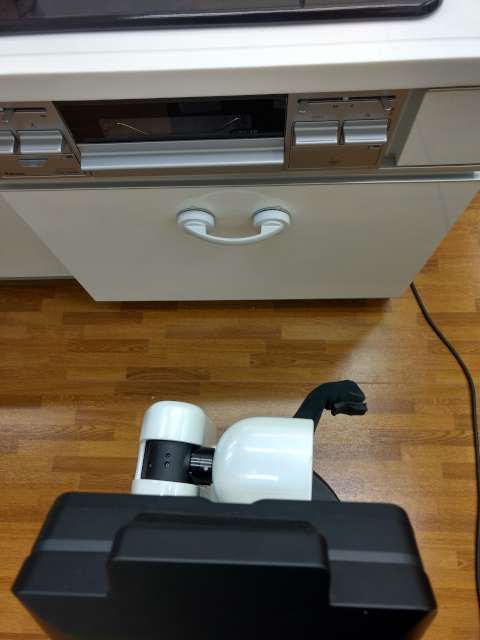}}
  		\caption{}
  		\label{fig:robottop}
	\end{subfigure}
	\caption{a) The robot's distance to the dishwasher should allow manipulation without collisions; b) the upper RGBD camera is used to detect the dishwasher and door handle positions with respect to the gripper in the context of object manipulation.}
	\label{fig:visualfeedback}
\end{figure}

We propose an active object manipulation systems using a 3-DOF RGBD camera (height, pan and tilt movements) on top of a service robot and a 6-axis force sensor in the hand. Through this sensors, the robot is able to detect the obstacle's position and orientation in robot coordinates while the different states of the manipulation process take place (Figure \ref{fig:visualfeedback}). 

In particular, the robot arrives near the dishwasher within an uncertainty given by the localisation system based on 2D laser scans, but with a localisation error big enough to affect the performance in the dishwasher's door handle grasping step using only the arm's inverse kinematics. Therefore, we propose the use of the upper RGBD camera to update the robot's relative position to the dishwasher and to locate the handle, and then we use the contact sensor in the robot manipulator to detect when the robot reaches it. This approach allowed us to be the only team in the category able to open the dishwasher door at the RoboCup2018.

\subsection{Active vision-based system}

\begin{figure}[h]
	\centering
	\includegraphics[width=0.45\textwidth]{{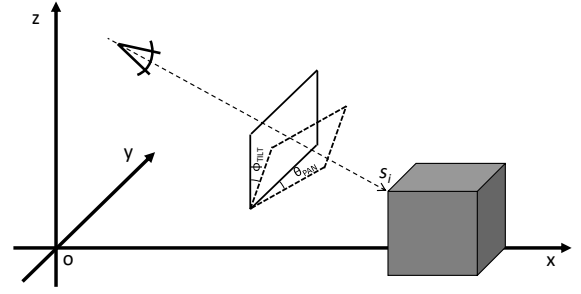}}
	\caption{From a 3-DOF RGBD camera we obtain a point cloud of 3D features in robot coordinates after applying a series of geometric transformations.}
	\label{fig:transformation}
\end{figure}

We start with a point cloud of 3D features from the RGBD camera on top of the robot. We transform the camera's coordinates system to a robot coordinate system, as shown in Figure \ref{fig:transformation}, by applying a series of geometric transformations as follows. Given a 3D point $s_i = (x_i, y_i, z_i)$ in camera coordinates, and a known camera orientation, $\phi_{TILT}$ and $\theta_{PAN}$ and position $t$ we obtain the point in robot coordinates as

$$p_i = Ps_i = \begin{bmatrix} R & t \\ 0 & 1 \end{bmatrix} s_i$$

where

$$ R = R_{Y}R_{Z} =
\begin{bmatrix}
\cos(\phi) & 0 & \sin(\phi) \\
0 & 1 & 0 \\
-\sin(\phi) & 0 & \cos(\phi)
\end{bmatrix}
\begin{bmatrix}
\cos(\theta) & -\sin(\theta) & 0 \\
\sin(\theta) & \cos(\theta) & 0 \\
0 & 0 & 1
\end{bmatrix}
$$

and

$$t = 
\begin{bmatrix}
t_x & t_y & t_z
\end{bmatrix}^T
$$

\subsubsection{Robot pose correction}

Given a point cloud in robot coordinates, we project the XYZ point cloud to the horizontal plane XY and find the set of points $S$ in the robot's line-of-sight, and extract the straight line $l$ that best fit this set using RANSAC \cite{fischler:1981}. The minimum number of iterations can be found as 

$$N = \frac{\log(1 - p)}{\log(1 - (1 - \epsilon )^s)}$$

where $p$ is the probability of success, $\epsilon$ the percentage of outliers in the dataset, and $s$ is the size of $S$. The distance from a point $q$ to a line defined by two points, $p_1$ and $p_2$, is calculated as follows

$$d(p_1, p_2, q) = \frac{|(q-p_1)\times(q-p_2)|}{|p_2-p_1|} $$

Finally, we obtain the normal vector to the robot's frontal-plane 

$$n = (n_x, n_y, n_z) = (1, 0, 0)$$

and calculate the angle $\omega$ between the edge $l$ and normal vector $n$

$$\cos(\omega) = \frac{l \cdot n}{|l| |n|}$$

and the distance to the edge as the average $\hat{l}_x$ component of the extreme points in $l$ after orientation correction, i.e. $\hat{l}$, defined as

$$\hat{l} = R_{Z}l = 
\begin{bmatrix}
\cos(\omega) & -\sin(\omega) & 0 \\
\sin(\omega) & \cos(\omega) & 0 \\
0 & 0 & 1
\end{bmatrix}
\begin{bmatrix}
l_x \\
l_y \\
l_z
\end{bmatrix}$$

\begin{figure}[ht]
\centering
	\begin{subfigure}{.225\textwidth}
  		\centering
  		\includegraphics[width=1\textwidth]{{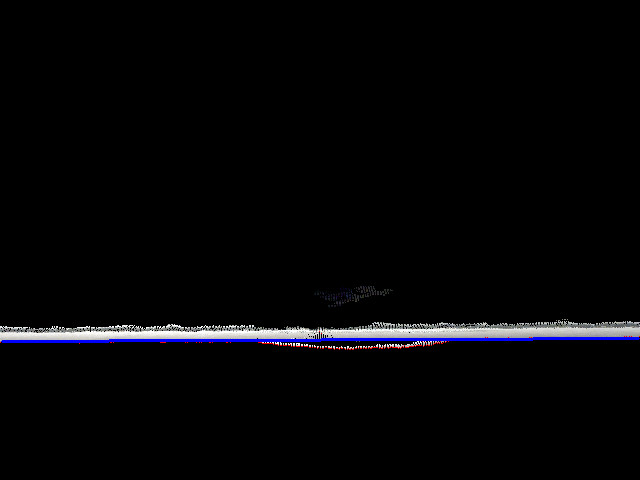}}
  		\caption{}
  		\label{fig:linepovgood}
	\end{subfigure}
	\begin{subfigure}{.225\textwidth}
  		\centering
  		\includegraphics[width=1\textwidth]{{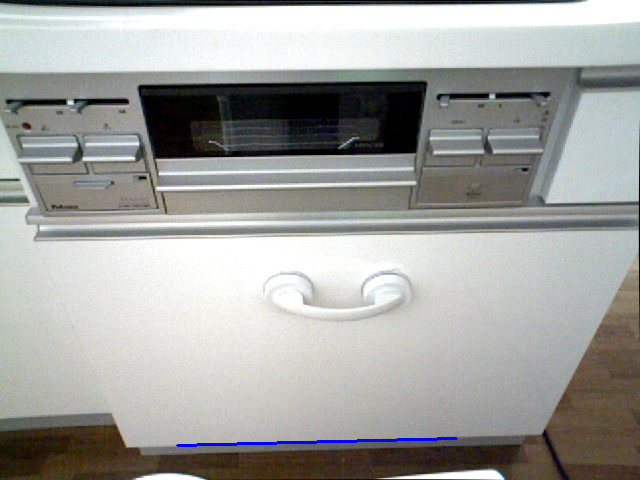}}
  		\caption{}
  		\label{fig:linergbgood}
	\end{subfigure}
	\caption{Edge finder a) top (metric) and b) front (RGB) view output for the dishwasher scene.}
	\label{fig:edgegoodpose} 
\end{figure}

In Figure \ref{fig:linepovgood}, it can be observed the result in robot coordinates after performing RANSAC in the XY plane -- the input point cloud was limited in depth, height, and width so it doesn't include the floor plane nor objects to far from the robot. In Figure \ref{fig:linergbgood} we show the found edge in camera coordinates.

\subsubsection{Object position detection}

\begin{figure}[ht]
\centering
	\begin{subfigure}{.225\textwidth}
  		\centering
  		\includegraphics[width=1\textwidth]{{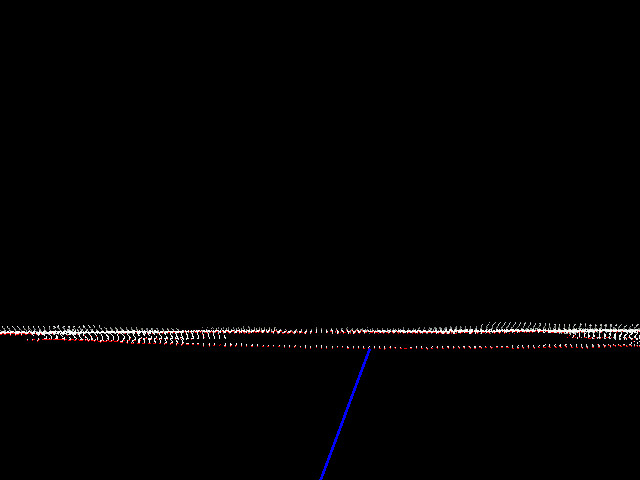}}
  		\caption{}
  		\label{fig:pointpovgood}
	\end{subfigure}
	\begin{subfigure}{.225\textwidth}
  		\centering
  		\includegraphics[width=1\textwidth]{{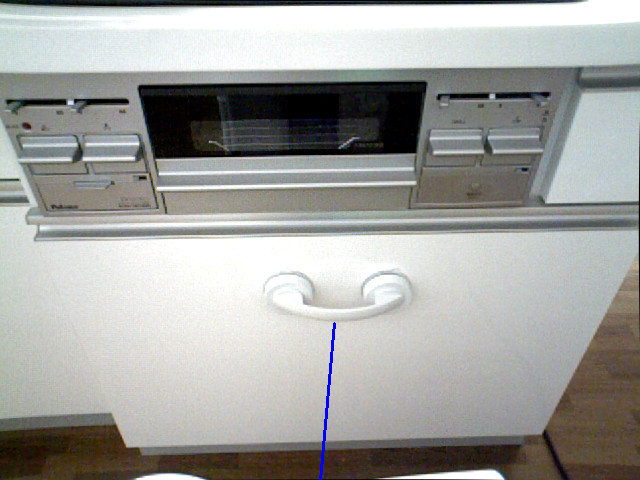}}
  		\caption{}
  		\label{fig:pointrgbgood}
	\end{subfigure}
	\caption{Door handle a) top (metric) and b) front (RGB) detection.}
	\label{fig:handlegoodpose} 
\end{figure}

Similarly, given a point cloud in robot coordinates and the robot being aligned to the dishwasher, from the plane XY view we find the closest point $p_c$ to the camera plane, i.e. the point $p_c$ with the lowest $x$ value, as shown in Figure \ref{fig:handlegoodpose}.

\subsection{Active contact-based system}

\begin{figure}[ht]
\centering
	\begin{subfigure}{.1500\textwidth}
  		\centering
  		\includegraphics[width=1\textwidth]{{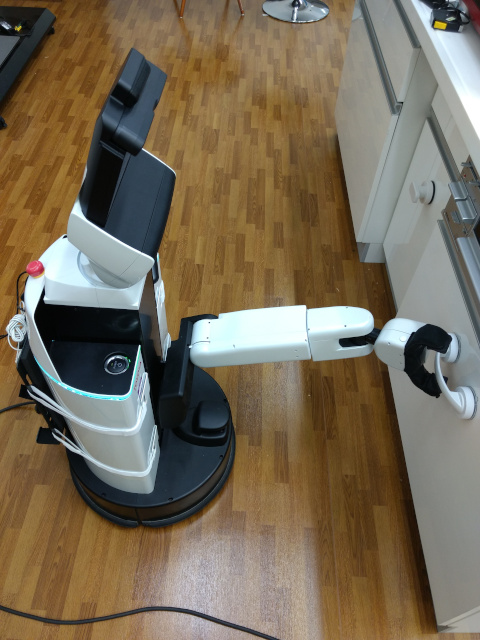}}
  		\label{fig:robotdoor01a}
	\end{subfigure}
	\begin{subfigure}{.1500\textwidth}
  		\centering
  		\includegraphics[width=1\textwidth]{{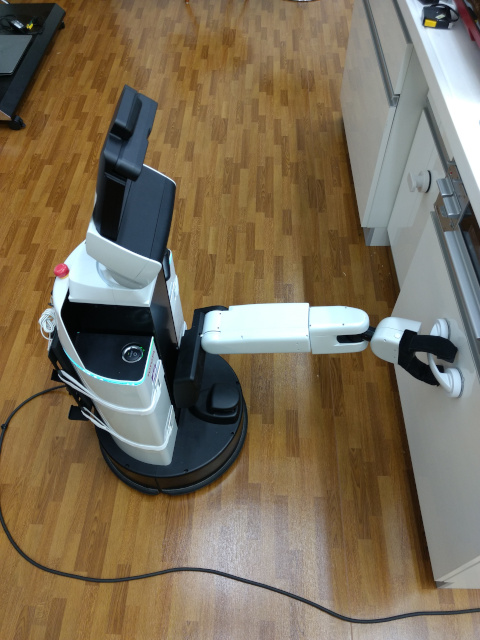}}
  		\label{fig:robotdoor02a}
	\end{subfigure} 
	\begin{subfigure}{.1500\textwidth}
  		\centering
  		\includegraphics[width=1\textwidth]{{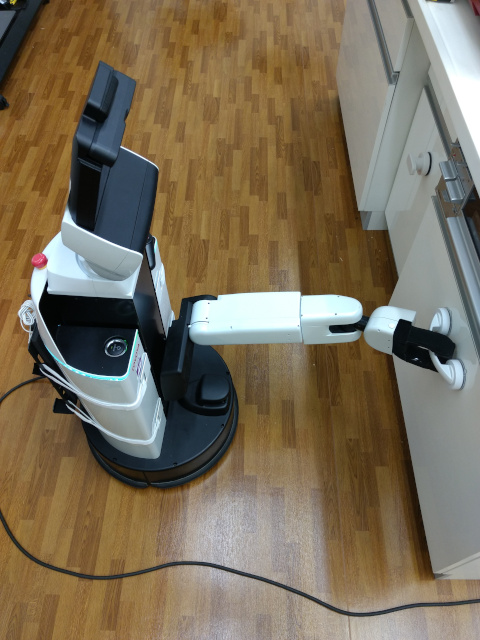}}
  		\label{fig:robotdoor03a}
	\end{subfigure} \\
	\begin{subfigure}{.1500\textwidth}
  		\centering
  		\includegraphics[width=1\textwidth]{{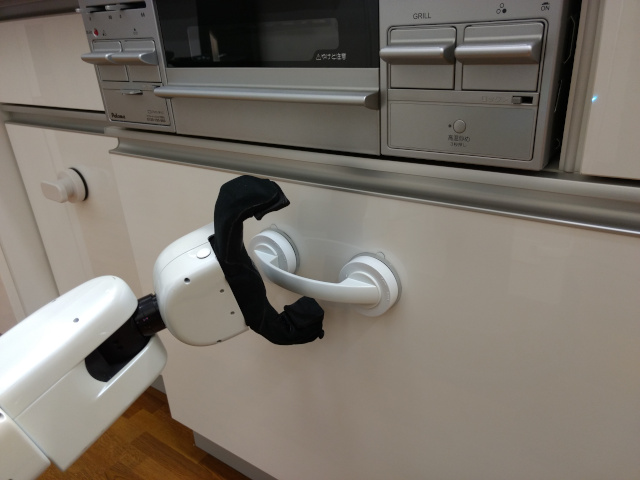}}
  		\caption{}
  		\label{fig:robotdoor01b}
	\end{subfigure}
	\begin{subfigure}{.1500\textwidth}
  		\centering
  		\includegraphics[width=1\textwidth]{{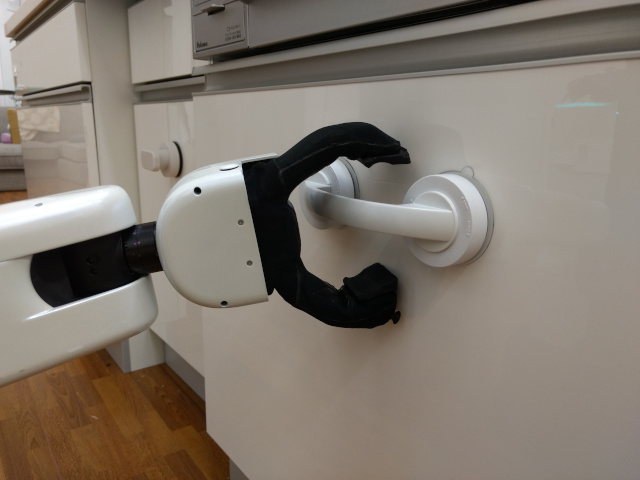}}
  		\caption{}
  		\label{fig:robotdoor02b}
	\end{subfigure}
	\begin{subfigure}{.1500\textwidth}
  		\centering
  		\includegraphics[width=1\textwidth]{{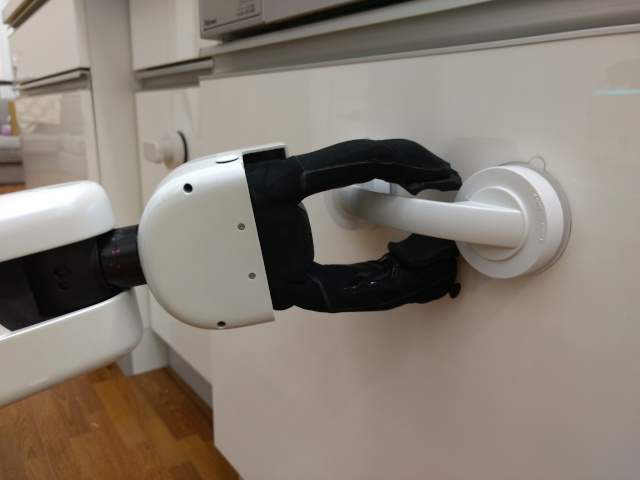}}
  		\caption{}
  		\label{fig:robotdoor03b}
	\end{subfigure} 
	\caption{Gripper to door handle approximation and contact sensor usage.}
	\label{fig:robotdoor}
\end{figure}

The Toyota HSR is able to detect a force applied to the hand by a series of 6 axis force sensors mounted on the wrist. We align the hand's X force axis to a normal vector to the obstacle's plane and then we move the gripper in the direction of the target object until we detect that the manipulator touches the obstacle's plane with the force sensor to finally close it to grasp the object. Figure \ref{fig:robotdoor} shows the result of this approach where the obstacle corresponds to the dishwasher and the target object to the door handle.

\section{Experiments and Results} \label{sect:experiments}

We present our results on multimodal feedback for object manipulation, namely, relative robot position and orientation correction with respect to obstacles edges and object position. In the experiments, for RANSAC we set $p = 99\%$, $\epsilon = 50\%$, and MIN = 0.01m, and performed the dishwasher door opening. 

\begin{figure}[ht]
	\centering
	\begin{subfigure}{.225\textwidth}
  		\centering
  		\includegraphics[width=1\textwidth]{{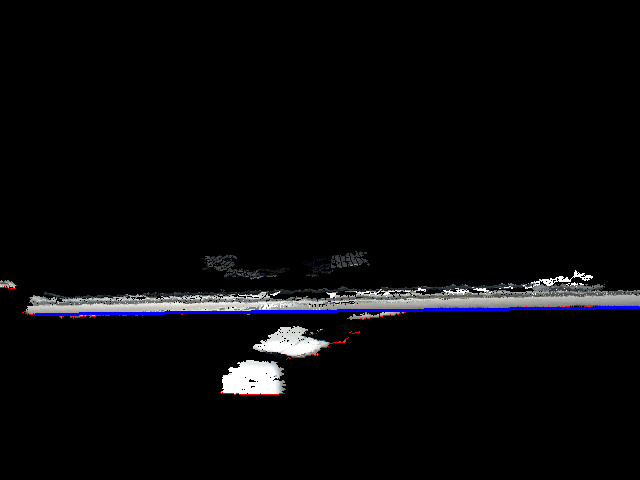}}
  		\caption{}
  		\label{fig:linepovbad}
	\end{subfigure}
	\begin{subfigure}{.225\textwidth}
  		\centering
  		\includegraphics[width=1\textwidth]{{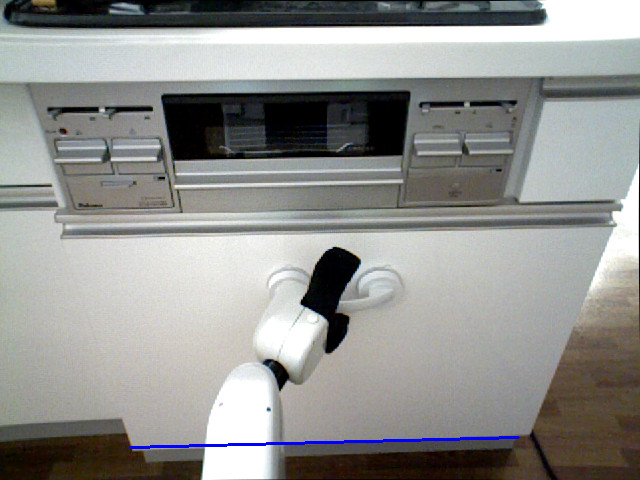}}
  		\caption{}
  		\label{fig:linergbbad}
	\end{subfigure} \\
	\begin{subfigure}{.225\textwidth}
  		\centering
  		\includegraphics[width=1\textwidth]{{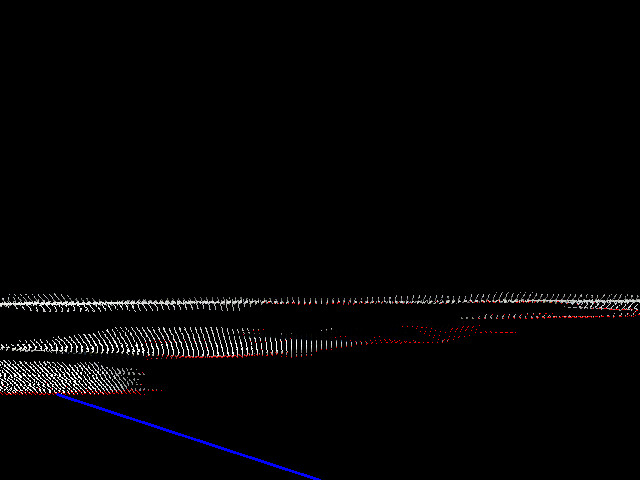}}
  		\caption{}
  		\label{fig:pointpovbad}
	\end{subfigure}
	\begin{subfigure}{.225\textwidth}
  		\centering
  		\includegraphics[width=1\textwidth]{{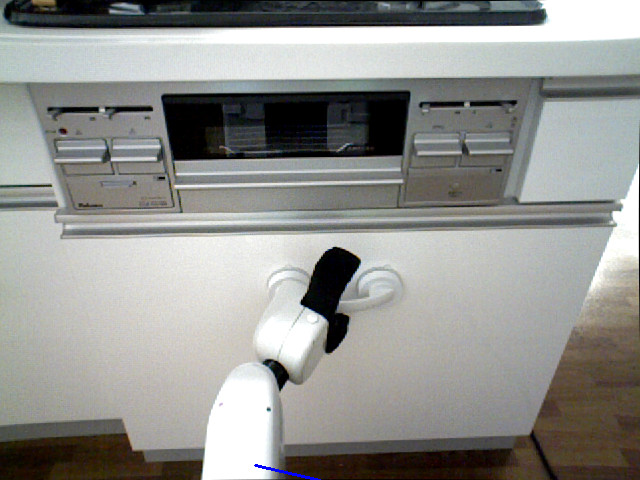}}
  		\caption{}
  		\label{fig:pointrgbbad}
	\end{subfigure}
	\caption{Edge finder a) top (metric) and b) front (RGB) view output in a table scene, and door handle c) top (metric) and d) front (RGB) detection. It can be observed that, in case of occlusions due to the manipulator, even though the obstacle edge alignment is still possible, the target object detection fails.}
	\label{fig:badpose} 
\end{figure}

The algorithm first align the robot to the dishwasher and then finds the door handle. While the robot manipulator approaches the target object, it occludes the top camera's view (Figure \ref{fig:linepovbad} and Figure \ref{fig:linergbbad}). It can be observed that this approach is robust to occlusions as long as segments of the main line are still visible, and therefore we know the relative position of the robot with respect to the obstacle edge, so we can adjust its pose accordingly to perform the manipulation task. However, the door handle occlusion make it unable to continue using visual feedback in the grasping step, as shown in Figure \ref{fig:pointpovbad} and Figure \ref{fig:pointrgbbad} and, in consequence, the contact sensor information is used to detect when the robot reaches the target to successfully performing the task. Finally, we use visual feedback to know whether the door was opened successfully.

In the proposed task, the robot starts at a known position and has to arrive in front of the dishwasher and aim to open it. To fairly evaluate the performance using only mechanical information (i.e. no visual feedback), the experimental design included two main configurations: one where the robot position in front of the dishwasher is constant -- i.e. hard-coded and therefore it required frequent and time-consuming map initialisation to reduce to the minimum the cumulated drift error -- and one where the robot arrives around a known position with a random drift in position and orientation as follows. The lateral drift was a uniform random number between -0.1 and 0.1 meters while the frontal drift was between -0.2 and 0.0 meters, i.e. positions closer to the dishwasher where omitted to avoid collisions due to the lack of visual feedback in some methods. The orientation drift went from -15 to 15 degrees, and the door handle position randomly varied around its center in $x$ and $y$ with values from -0.05 to 0.05 meters.

We performed a series of experiments to evaluate our approach in the door opening task and compare it with a number of baselines as follows. First, with no-feedback more than the arm's inverse kinematics and a method that only uses tactile feedback, where the robot arrives to a known position in front of the dishwasher and executes a series of deterministic state machines. Then, we evaluated a method that only uses visual information and a method that uses both visual and tactile information. We run 10 experiments per method (feedback input) per configuration (constant or random position in front of the drawer). Results are reported in Table \ref{tab:feedback}.

\begin{table}[htp]
\centering
	\caption{Door-opening average performance in a known (constant) and unknown (random) setup after ten trials per method.}
	\label{tab:feedback}
	\centering
	\resizebox{0.45\textwidth}{!}{
		\begin{tabular}{|c|c|c|c|c|}
		\hline
			&\multicolumn{4}{|c|}{Manipulation feedback}\\ 
		\cline{2-5}
			Pose&No-Feedback&Tactile&Visual&Tactile+Visual\\
		\hline
			Constant & 0.7 & 0.6 & 1.0 & 1.0\\
			Random & 0.2 & 0.2 &  1.0 & 1.0 \\
		\hline
		\end{tabular}}
\end{table}

It can be observed that, while methods that only use mechanical feedback perform well in known setups, adding visual information increases their performance, even in unknown configurations. Although using visual information seems to be enough, tactile information is useful to confirm the manipulator pose, or in the opposite problem, where the robot has to close a door: the robot should move until it detects an increase in the force in the motion axis. An overall performance in the dishwasher's door opening task can be seen at https://youtu.be/zuLe1wamo8A . This approach was successfully tested in the RoboCup2018 Procter and Gamble (TM) Dishwasher Challenge. 

\section{Conclusions}

In this work we present our results on multimodal active robot-object interaction using laser, visual and contact feedback. In the manipulation task, the robot workspace is restricted by the arm's configuration space so an active robot and object localisation is desired in dynamic environment such that the arm's configuration space intersects with the target object's position. After standard laser-based localisation, we introduced our edge extraction module given a point cloud of 3D features from an RGBD camera on top of the robot, and used it to correct the robot pose (position and orientation) with respect to the obstacle, and also use visual and contact feedback in the grasping task so the robot can actively adjust the gripper pose with respect to the object to detect changes in the target's position or whether the robot has reached it, and to confirm if the task has been executed successfully. We compare our method with a series of baselines obtaining better performance. The aim of this work is not to present the state-of-the-art on a given problem but to start the discussion on how manipulation intelligence can be achieved through active reasoning, where we favor the integration of several multi-task systems instead of a task-specific solution.

\addtolength{\textheight}{-15cm} 





\section*{ACKNOWLEDGMENT}
This work was supported by CREST, JST.



\bibliographystyle{ieeetr}
\bibliography{all}

\begin{thebibliography}{10}

\bibitem{wisspeintner:2009}
T.~Wisspeintner, T.~Van Der~Zant, L.~Iocchi, and S.~Schiffer, ``Robocup@ home:
  Scientific competition and benchmarking for domestic service robots,'' {\em
  Interaction Studies}, vol.~10, no.~3, pp.~392--426, 2009.

\bibitem{amigoni:2015}
F.~Amigoni, E.~Bastianelli, J.~Berghofer, A.~Bonarini, G.~Fontana,
  N.~Hochgeschwender, L.~Iocchi, G.~Kraetzschmar, P.~Lima, M.~Matteucci, {\em
  et~al.}, ``Competitions for benchmarking: task and functionality scoring
  complete performance assessment,'' {\em IEEE Robotics \& Automation
  Magazine}, vol.~22, no.~3, pp.~53--61, 2015.

\bibitem{contreras:2017}
L.~Contreras and W.~Mayol-Cuevas, ``{O-POCO: Online POint cloud COmpression
  mapping for visual odometry and SLAM},'' in {\em {International Conference on
  Robotics and Automation}}, {IEEE}, 2017.

\bibitem{savage:2018}
J.~Savage, O.~Fuentes, L.~Contreras, and M.~Negrete, ``{Map representation
  using hidden markov models for mobile robot localization},'' in {\em {13th
  International Scientific-Technical Conference on Electromechanics and
  Robotics}}, 2018.

\bibitem{potthast:2016}
C.~Potthast, A.~Breitenmoser, F.~Sha, and G.~S. Sukhatme, ``Active multi-view
  object recognition: A unifying view on online feature selection and view
  planning,'' {\em Robotics and Automation Systems}, vol.~84, pp.~31--47, 2016.

\bibitem{cadena:2016}
C.~Cadena, L.~Carlone, H.~Carrillo, Y.~Latif, D.~Scaramuzza, J.~Neira, I.~Reid,
  and J.~J. Leonard, ``Past, present, and future of simultaneous localization
  and mapping: Toward the robust-perception age,'' {\em IEEE Transactions on
  Robotics}, vol.~32, no.~6, pp.~1309--1332, 2016.

\bibitem{bajcsy:2018}
R.~Bajcsy, Y.~Aloimonos, and J.~K. Tsotsos, ``Revisiting active perception,''
  {\em Autonomous Robots}, vol.~42, no.~2, pp.~177--196, 2018.

\bibitem{hashimoto:2013}
K.~Hashimoto, F.~Saito, T.~Yamamoto, and K.~Ikeda, ``A field study of the human
  support robot in the home environment,'' in {\em 2013 IEEE Workshop on
  Advanced Robotics and its Social Impacts}, pp.~143--150, 2013.

\bibitem{quigley:2009}
M.~Quigley, K.~Conley, B.~Gerkey, J.~Faust, T.~Foote, J.~Leibs, R.~Wheeler, and
  A.~Y. Ng, ``Ros: an open-source robot operating system,'' in {\em ICRA
  workshop on open source software}, vol.~3, p.~5, 2009.

\bibitem{matamoros:2018}
M.~Matamoros, C.~Rascon, J.~Hart, D.~Holz, and L.~van Beek, ``Robocup@home
  2018: Rules and regulations.''
  \url{http://www.robocupathome.org/rules/2018_rulebook.pdf}, 2018.

\bibitem{fischler:1981}
M.~A. Fischler and R.~C. Bolles, ``Random sample consensus: A paradigm for
  model fitting with applications to image analysis and automated
  cartography,'' {\em Communications of the ACM}, vol.~24, no.~6, pp.~381--395,
  1981.

\end{thebibliography}

\end{document}